\documentclass[11pt]{article}
\usepackage{acl2016}
\usepackage{times}
\usepackage{latexsym}
\usepackage{times}
\usepackage{url}
\usepackage{latexsym}
\usepackage[dvips]{graphicx}
\usepackage[table]{xcolor}
\usepackage{multirow}
\usepackage{xstring}
\usepackage{epstopdf}

\def\namecite{\newcite}

\aclfinalcopy 
\def\aclpaperid{193} 

\definecolor{light-green}{RGB}{178,255,102}
\definecolor{light-red}{RGB}{255,153,153}

\newcommand{\significant}[2]{
			\IfEqCase{#2}{%
			{+}{\cellcolor{light-green}#1$^{#2}$}%
			{++}{\cellcolor{green}#1$^{#2}$}%
			{-}{\cellcolor{light-red}#1$^{#2}$}
			{--}{\cellcolor{red}#1$^{#2}$}
    }
    [\PackageError{\significant}{Undefined option to tree: #1}{}]
    }

\title{One model, two languages: training bilingual parsers with harmonized treebanks}

\author{David Vilares, Carlos G\'{o}mez-Rodr\'{\i}guez and Miguel A. Alonso \\
  Grupo LyS, Departamento de Computaci\'{o}n, Universidade da Coru\~{n}a \\
  Campus de A Coru\~{n}a s/n, 15071, A Coru\~{n}a, Spain \\
  {\tt \{david.vilares, carlos.gomez, miguel.alonso\}@udc.es} 
  }
\date{}

\begin{document}

\maketitle

\begin{abstract}

We introduce an approach to train lexicalized parsers using bilingual corpora obtained by merging harmonized treebanks of different languages, producing parsers that can analyze sentences in either of the learned languages, or even sentences that mix both. 
We test the approach on the Universal Dependency Treebanks, training with MaltParser and MaltOptimizer.
The results show that these bilingual parsers are more than competitive,
as most combinations not only preserve accuracy, but some even achieve significant improvements over the corresponding monolingual parsers. 
Preliminary experiments also show the approach to be promising on texts with code-switching and when more languages are added.
\end{abstract}

\section{Introduction}

The need of frameworks for analyzing content 
in different languages has been discussed recently \cite{DanZhaJenBroKuWanChe2014a},
and multilingual dependency parsing is no stranger to this challenge.
Data-driven parsing models \cite{NivreTSF06} can be trained for any
language, given enough annotated data.

On languages where treebanks are not available, cross-lingual transfer can be used to train parsers for a target
language with data from one or more source languages. Data transfer approaches (e.g. \namecite{YarNgaWic2001}, \namecite{Tiedemann2014})
map linguistic annotations across languages through parallel corpora. 
Instead, model transfer approaches (e.g. \namecite{NasBarGlob2012}) rely 
on cross-linguistic syntactic regularities to learn aspects of the source language 
that help parse an unseen language, without parallel corpora.

Model transfer approaches have benefitted from the development of multilingual resources that harmonize annotations. \namecite{PetDasMcD2011a} proposed a universal tagset, and \namecite{McDPetHal2011} employed it to transfer delexicalized parsers \cite{ZenRes06}.
More recently, several projects have presented treebank collections of multiple languages with their annotations standardized at the syntactic level, including HamleDT \cite{ZemMarPopRamSteZabHaj2012a} and the Universal Dependency Treebanks \cite{McdNivQuirGolDasGanHalPetZhaTacBedCasLee2013}.

In this paper we also rely on these resources, but with a different goal:
we use universal annotations to
train bilingual dependency parsers that effectively analyze unseen sentences 
in any of the learned languages. Unlike delexicalized approaches for model transfer,
our parsers exploit lexical features.
The results are encouraging: our experiments show that, starting with a monolingual parser, 
we can ``teach'' it an additional language for free in terms of accuracy (i.e., without significant accuracy loss on the original language, in spite of learning a more complex task) in the vast majority of cases.

\section{Bilingual training}

Universal Dependency Treebanks v2.0 \cite{McdNivQuirGolDasGanHalPetZhaTacBedCasLee2013} is
a set of CoNLL-formatted treebanks for ten languages, annotated with common criteria. 
They include two versions of PoS tags: universal tags \cite{PetDasMcD2011a} 
in the {\sc cpostag} column, and a refined annotation with treebank-specific information 
in the {\sc postag} column. Some of the latter tags are not part of the core universal set, and
they can denote 
linguistic phenomena that are language-specific, or phenomena that not
all the corpora have annotated in the same way.

To train monolingual parsers (our baseline), we used the official training-dev-set splits provided 
with the corpora. For the bilingual models, for each pair of languages $L_1, L_2$; we simply 
merged their training sets into a single file acting as a training set for $L_1 \cup L_2$, and we did 
the same for the development sets. The test sets were not merged because comparing the 
bilingual parsers to monolingual ones requires evaluating each bilingual parser on the 
two corresponding monolingual test sets.

To build the models, we relied on MaltParser \cite{NivHalNilChaEryKubMarMar2007a}.
Due to the large number of language pairs that complicates manual optimization, and 
to ensure a fair comparison,
we used Malt\-Optimizer \cite{BalNiv2012}, 
an automatic optimizer for
MaltParser models. This system works in three phases: 
\emph{Phase 1} and \emph{2} choose a parsing algorithm by analyzing the training set,
and performing experiments with default features.
\emph{Phase 3} tunes the feature model and algorithm parameters. We hypothesize that the bilingual models will learn a set of features that fits both languages, and check this hypothesis by evaluating on the test sets.

We propose two training configurations:
(1) a \emph{treebank-dependent tags} configuration
where we include the information in the {\sc postag} column and (2) 
a \emph{universal tags only} configuration,
where we do not use this information, relying only on the {\sc cpostag} column.
Information that could
be present in {\sc feats} or {\sc lemma} columns is not used in any case. 
This methodology plans to answer two research questions:
(1) can we train bilingual parsers with good accuracy by
merging harmonized training sets?, and (2) is it essential that the tagsets for both languages
are the same, or can we still get accuracy gains from fine-grained PoS tags (as in the monolingual case)
even if some of them are treebank-specific? 

All models are freely available.\footnote{\url{http://grupolys.org/software/PARSERS/}}

\section{Evaluation}

\begin{table*}[t]
\begin{center}
\scriptsize{
 \tabcolsep=0.075cm
\begin{tabular}{|l|llllllllll|}
\hline 
&\bf\emph{de}&\bf\emph{en}&\bf\emph{es}
&\bf\emph{fr}&\bf\emph{id}&\bf\emph{it}
&\bf\emph{ja}&\bf\emph{ko}&\bf\emph{pt-br}
&\bf\emph{sv}\\
\hline
\multirow{2}{*}{\bf\emph{de}}&78.27&\significant{78.01}{-}&\significant{77.82}{-}&\significant{77.83}{-}&\significant{77.84}{-}&\significant{78.10}{-}&\significant{77.86}{-}&\significant{77.94}{-}&\significant{78.13}{-}&\significant{78.60}{+}\\
	                     &84.03&\significant{84.08}{+}&\significant{83.82}{-}&\significant{83.55}{--}&\significant{83.85}{-}&\significant{84.12}{+}&\significant{83.88}{-}&\significant{83.63}{-}&\significant{83.87}{-}&\significant{84.38}{+}\\			      
\hline
 \multirow{2}{*}{\bf\emph{en}}&\significant{89.37}{+}&89.36&\significant{89.46}{+}&\significant{89.38}{+}&\significant{89.69}{++}&\significant{89.82}{++}&\significant{89.43}{+}&\significant{89.63}{++}&\significant{89.60}{++}&\significant{89.11}{--}\\
 	                      &\significant{91.02}{+}&91.02&\significant{91.09}{+}&\significant{91.06}{+}&\significant{91.32}{++}&\significant{91.47}{++}&\significant{91.10}{+}&\significant{91.32}{++}&\significant{91.24}{+}&\significant{90.79}{--}\\			      
\hline
\multirow{2}{*}{\bf\emph{es}}&\significant{80.85}{+}&\significant{81.08}{++}&80.60&\significant{80.95}{+}&\significant{81.16}{+}&\significant{80.92}{+}&\significant{81.41}{++}&\significant{81.49}{++}&\significant{79.96}{-}&\significant{81.26}{++}\\
	                     &\significant{85.17}{+}&\significant{85.27}{++}&84.75&\significant{85.15}{+}&\significant{85.00}{+}&\significant{85.13}{+}&\significant{85.52}{++}&\significant{85.39}{++}&\significant{84.70}{-}&\significant{85.42}{++}\\			      
 \hline
 \multirow{2}{*}{\bf\emph{fr}}&\significant{79.01}{-}&\significant{79.39}{+}&\significant{79.36}{+}&79.29&\significant{79.61}{+}&\significant{79.34}{+}&\significant{79.16}{-}&\significant{79.36}{+}&\significant{79.09}{-}&\significant{79.66}{+}\\
 	                      &\significant{84.17}{-}&\significant{84.49}{+}&\significant{84.56}{+}&84.47&\significant{84.32}{-}&\significant{84.41}{-}&\significant{84.34}{-}&\significant{84.72}{+}&\significant{83.98}{-}&\significant{84.84}{+}\\			      
\hline
\multirow{2}{*}{\bf\emph{id}}&\significant{75.72}{--}&\significant{77.19}{-}&\significant{77.12}{-}&\significant{77.15}{-}&77.69&\significant{78.29}{+}&\significant{77.60}{-}&\significant{76.68}{--}&\significant{77.45}{-}&\significant{77.01}{--}\\
	                     &\significant{81.73}{--}&\significant{82.66}{--}&\significant{82.72}{-}&\significant{82.66}{-}&83.38&\significant{84.09}{+}&\significant{83.18}{-}&\significant{82.16}{--}&\significant{82.96}{-}&\significant{82.59}{--}\\			                           
\hline
\multirow{2}{*}{\bf\emph{it}}&\significant{82.62}{--}&\significant{83.17}{--}&\significant{83.12}{--}&\significant{83.10}{--}&\significant{83.74}{--}&84.40&\significant{84.62}{+}&\significant{84.79}{+}&\significant{83.70}{-}&\significant{84.55}{+}\\
 	                     &\significant{86.14}{--}&\significant{86.46}{--}&\significant{86.78}{-}&\significant{86.69}{-}&\significant{86.73}{--}&87.54&\significant{87.48}{-}&\significant{87.46}{-}&\significant{87.39}{-}&\significant{87.23}{-}\\			      
\hline
\multirow{2}{*}{\bf\emph{ja}}&\significant{76.53}{--}&\significant{76.24}{--}&\significant{76.61}{--}&\significant{76.32}{--}&\significant{75.18}{--}&\significant{77.05}{-}&77.46&\significant{76.89}{-}&\significant{76.69}{-}&\significant{76.89}{-}\\
 	                     &\significant{83.77}{-}&\significant{83.89}{-}&\significant{84.26}{-}&\significant{84.05}{--}&\significant{83.08}{--}&\significant{83.97}{-}&84.34&\significant{83.65}{-}&\significant{83.97}{-}&\significant{84.17}{-}\\			      
\hline
\multirow{2}{*}{\bf\emph{ko}}&\significant{86.13}{--}&\significant{88.30}{+}&\significant{87.91}{+}&\significant{88.49}{+}&\significant{85.86}{--}&\significant{88.72}{++}&\significant{87.14}{--}&87.83&\significant{86.75}{--}&\significant{88.68}{-}\\
	                     &\significant{90.61}{--}&\significant{92.16}{+}&\significant{92.00}{-}&\significant{92.35}{+}&\significant{90.19}{--}&\significant{92.55}{+}&\significant{91.89}{-}&92.12&\significant{91.39}{--}&\significant{92.39}{-}\\			      
 \hline
 \multirow{2}{*}{\bf\emph{pt-br}}&\significant{84.83}{-}&\significant{85.06}{+}&\significant{84.99}{+}&\significant{84.97}{+}&\significant{85.10}{+}&\significant{85.43}{++}&\significant{84.95}{+}&\significant{85.12}{+}&84.88&\significant{85.25}{++}\\
 	                     &\significant{87.18}{-}&\significant{87.19}{-}&\significant{87.27}{+}&\significant{87.17}{-}&\significant{87.35}{-}&\significant{87.68}{++}&\significant{87.13}{-}&\significant{87.35}{-}&87.39&\significant{87.43}{++}\\			      
\hline
 \multirow{2}{*}{\bf\emph{sv}}&\significant{81.71}{--}&\significant{82.01}{--}&\significant{82.03}{-}&\significant{81.92}{--}&\significant{82.34}{-}&\significant{82.63}{+}&\significant{82.81}{+}&\significant{82.94}{++}&\significant{82.19}{-}&82.48\\
 	                     &\significant{86.01}{--}&\significant{86.39}{-}&\significant{86.55}{-}&\significant{86.28}{--}&\significant{86.69}{-}&\significant{86.55}{-}&\significant{86.92}{+}&\significant{86.83}{-}&\significant{86.39}{-}&86.92\\			                        
 	                    
\hline
\end{tabular}}
\end{center}
\caption{\small{Performance on the Universal Dependency Treebanks test sets using the gold {\sc postag} information. For each cell, its (row,column) pair indicates the language(s) with which the model was trained, with the row corresponding to the language where it was evaluated. 
\emph{\textquoteleft \colorbox{green}{++}\textquoteright} and \emph{\textquoteleft \colorbox{light-green}{+}\textquoteright} indicate
that the improvement in performance obtained by the bilingual model is statistically significant or not, respectively.
\emph{\textquoteleft \colorbox{red}{-\ -}\textquoteright} and \emph{\textquoteleft \colorbox{light-red}{-}\textquoteright} correspond to significant and not significant \emph{decreases} in accuracy.}}
\label{table-different-tags-sets-performance}
\end{table*}

\begin{table*}[t]
\begin{center}
\scriptsize{
 \tabcolsep=0.075cm
\begin{tabular}{|l|llllllllll|}
\hline 
&\bf\emph{de}&\bf\emph{en}&\bf\emph{es}
&\bf\emph{fr}&\bf\emph{id}&\bf\emph{it}
&\bf\emph{ja}&\bf\emph{ko}&\bf\emph{pt-br}
&\bf\emph{sv}\\
\hline
\multirow{2}{*}{\bf\emph{de}}&74.07&\significant{72.04}{--}&\significant{74.51}{+}&\significant{74.44}{+}&\significant{73.68}{-}&\significant{73.76}{-}&\significant{73.90}{-}&\significant{74.30}{+}&\significant{74.29}{+}&\significant{74.76}{++}\\
                             &79.77&\significant{77.52}{--}&\significant{79.95}{+}&\significant{79.83}{+}&\significant{79.24}{-}&\significant{79.44}{-}&\significant{79.83}{+}&\significant{79.76}{-}&\significant{79.71}{-}&\significant{80.25}{+}\\
\hline
 \multirow{2}{*}{\bf\emph{en}}&\significant{88.46}{+}&88.35&\significant{88.65}{++}&\significant{88.39}{+}&\significant{88.61}{++}&\significant{88.68}{++}&\significant{88.65}{++}&\significant{88.61}{++}&\significant{88.65}{++}&\significant{88.50}{+}\\
                              &\significant{90.35}{+}&90.27&\significant{90.54}{++}&\significant{90.26}{-}&\significant{90.47}{++}&\significant{90.53}{++}&\significant{90.49}{++}&\significant{90.43}{++}&\significant{90.55}{++}&\significant{90.43}{++}\\
 \hline
\multirow{2}{*}{\bf\emph{es}}&\significant{79.66}{--}&\significant{78.78}{--}&80.54$^{}$&\significant{79.59}{--}&\significant{78.98}{--}&\significant{79.84}{--}&\significant{79.59}{--}&\significant{79.80}{--}&\significant{79.74}{--}&\significant{79.09}{--}\\
                              &\significant{83.81}{--}&\significant{82.94}{--}&84.35$^{}$&\significant{83.26}{--}&\significant{82.79}{--}&\significant{83.79}{--}&\significant{83.53}{--}&\significant{83.57}{-}&\significant{83.76}{--}&\significant{83.28}{--}\\
 \hline
\multirow{2}{*}{\bf\emph{fr}}&\significant{78.43}{+}&\significant{78.10}{-}&\significant{78.63}{+}&78.40$^{}$&\significant{77.79}{-}&\significant{78.60}{+}&\significant{79.11}{+}&\significant{78.22}{-}&\significant{78.56}{+}&\significant{78.83}{+}\\
                               &\significant{83.26}{-}&\significant{82.77}{-}&\significant{83.38}{-}&83.40$^{}$&\significant{82.85}{-}&\significant{83.50}{+}&\significant{84.03}{+}&\significant{83.05}{-}&\significant{83.45}{+}&\significant{83.73}{+}\\
 \hline                             
 \multirow{2}{*}{\bf\emph{id}}&\significant{74.46}{--}&\significant{74.65}{--}&\significant{77.09}{--}&\significant{76.23}{--}&78.31$^{}$&\significant{77.86}{-}&\significant{77.10}{--}&\significant{75.58}{--}&\significant{76.90}{--}&\significant{78.34}{+}\\
                              &\significant{80.87}{--}&\significant{80.21}{--}&\significant{82.81}{--}&\significant{81.78}{--}&83.81$^{}$&\significant{83.52}{-}&\significant{82.68}{--}&\significant{81.20}{--}&\significant{82.50}{--}&\significant{83.83}{+}\\
 \hline
 \multirow{2}{*}{\bf\emph{it}}&\significant{82.27}{--}&\significant{82.13}{--}&\significant{82.24}{--}&\significant{82.75}{--}&\significant{82.65}{--}&83.88$^{}$&\significant{83.04}{--}&\significant{83.77}{-}&\significant{83.07}{--}&\significant{83.47}{-}\\
                              &\significant{85.40}{--}&\significant{85.38}{--}&\significant{85.36}{--}&\significant{86.31}{--}&\significant{85.45}{--}&86.68$^{}$&\significant{85.83}{--}&\significant{86.30}{-}&\significant{86.21}{--}&\significant{86.33}{-}\\
 \hline                             
 \multirow{2}{*}{\bf\emph{ja}}&\significant{69.41}{--}&\significant{68.88}{--}&\significant{69.28}{--}&\significant{69.24}{--}&\significant{69.73}{--}&\significant{70.22}{--}&70.87$^{}$&\significant{69.73}{--}&\significant{69.24}{--}&\significant{70.02}{-}\\
                              &\significant{79.62}{--}&\significant{79.21}{--}&\significant{79.45}{--}&\significant{80.11}{--}&\significant{79.58}{--}&\significant{79.58}{--}&81.16$^{}$&\significant{80.23}{-}&\significant{79.37}{--}&\significant{80.47}{--}\\
 \hline                            
 \multirow{2}{*}{\bf\emph{ko}}&\significant{84.40}{--}&\significant{84.82}{--}&\significant{85.40}{--}&\significant{84.59}{--}&\significant{84.74}{--}&\significant{86.79}{-}&\significant{86.21}{--}&87.52$^{}$&\significant{86.29}{--}&\significant{86.40}{--}\\
                              &\significant{89.61}{--}&\significant{90.00}{--}&\significant{90.77}{--}&\significant{89.88}{--}&\significant{90.00}{--}&\significant{91.39}{-}&\significant{91.46}{--}&92.00$^{}$&\significant{90.92}{--}&\significant{91.19}{--}\\
 \hline                           
 \multirow{2}{*}{\bf\emph{pt-br}}&\significant{83.40}{-}&\significant{82.76}{--}&\significant{83.56}{-}&\significant{83.72}{-}&\significant{83.08}{--}&\significant{83.95}{+}&\significant{83.80}{-}&\cellcolor{green}{84.16}{++}&83.83$^{}$&\cellcolor{green}{84.28}{++}\\
                                 &\significant{85.78}{-}&\significant{85.01}{--}&\significant{85.82}{-}&\significant{85.85}{-}&\significant{85.38}{--}&\significant{86.15}{+}&\significant{85.93}{-}&\significant{86.33}{+}&86.11$^{}$&\cellcolor{green}{86.41}{++}\\
  \hline                                
  \multirow{2}{*}{\bf\emph{sv}}&\significant{79.65}{--}&\significant{79.61}{--}&\significant{79.75}{--}&\significant{80.46}{-}&\significant{80.94}{+}&\significant{81.06}{+}&\significant{81.19}{+}&\significant{81.11}{+}&\significant{80.89}{-}&80.93$^{}$\\
                               &\significant{84.14}{--}&\significant{84.42}{--}&\significant{84.46}{--}&\significant{84.88}{-}&\significant{85.14}{-}&\significant{85.51}{+}&\significant{85.29}{-}&\significant{85.14}{-}&\significant{85.05}{-}&85.32$^{}$\\
 \hline
\end{tabular}}
\end{center}
\caption{\small{Performance on the Universal Dependency Treebanks test sets using the gold {\sc cpostag} information. The table is laid out like Table \ref{table-different-tags-sets-performance}}.
}
\label{table-same-tags-sets-performance}
\end{table*}

To ensure a fair comparison between monolingual and bilingual models, we chose to optimize the
models from scratch with MaltOptimizer, expecting it to choose the parsing algorithm and feature model which is
most likely to obtain good results.
We observed that the selection of a bilingual parsing algorithm was not necessarily related with 
the algorithms selected for the monolingual models. 
The system sometimes chose an algorithm for a bilingual model that was
not selected for any of the corresponding monolingual models.

In view of this, and as it is known that
different parsing algorithms can be more or less
competitive depending on the language \cite{Nivre2008}, we ran a control experiment to evaluate the models setting the same parsing algorithm for all cases, executing only 
\emph{phase 3} of MaltOptimizer. We chose the arc-eager parser for this experiment,
as it was the algorithm that MaltOptimizer chose most frequently for the monolingual models in the
previous configuration. The aim was to compare the accuracy of 
the bilingual models with respect to the monolingual ones, when there is no 
variation on the parsing algorithm between them. The results of this control experiment
are not shown for space reasons, but they were very similar to those of the original experiment.

\subsection{Results on the Universal Treebanks}\label{subsection-experiments-universal-treebanks}

Table \ref{table-different-tags-sets-performance} compares the accuracy of bilingual models
to that of monolingual ones, under the \emph{treebank-dependent tags} configuration. 
Each table cell shows the accuracy of a model, in terms of {\sc las} and {\sc uas}. Cells in the
diagonal correspond to monolingual models (the baseline), with the cell located at row $i$ and column $i$
representing the result obtained by training a monolingual parser on the training set of language $L_i$, and evaluating it on the test set of the same language $L_i$. Each cell outside the diagonal (at row $i$ and column $j$, with $j \neq i$) shows the results of training a bilingual model on the training set for $L_i \cup L_j$, evaluated on the test set of $L_i$.

As we can see, in a large majority of cases, bilingual parsers learn to parse two languages with
no statistically significant accuracy loss with respect to the corresponding monolingual parsers ($p<0.05$ with Bikel's randomized parsing evaluation comparator). This happened in 74 out of 90 cases when measuring {\sc uas}, or 69 out of 90 in terms of {\sc las}. Therefore, in most cases where
we are applying a parser to texts in a given language, adding a second language comes for free in terms of accuracy.

More strikingly, there are many cases where bilingual parsers outperform monolingual ones, even in this evaluation
on purely monolingual datasets. In particular, there are 12 cases where a bilingual parser obtains statistically
significant gains in {\sc las} over the monolingual baseline, and 9 cases with significant gains in {\sc uas}. 
This clearly surpasses the amount of significant gains to be expected by chance, and applying the Benjamini-Hochberg procedure \cite{benjamini1995controlling} to correct for multiple comparisons with a maximum false discovery rate of 20\% yields 8 significant improvements in {\sc las} and {\sc uas}. Therefore, it is clear that there is synergy between
datasets: in some cases, adding annotated data in a different language to our training set can actually improve the accuracy
that we obtain in the \emph{original} language. This opens up interesting research potential in using
confidence criteria to select the data that can help parsing in this way, akin to what is done in self-training approaches
\cite{Chen2008,GouAmb2011}.

Comparing the results by language, we note that the accuracy on the English and Spanish datasets almost always improves when adding a second treebank for training. Other languages that tend to get improvements in this way are French and Portuguese. There seems to be a rough trend towards the languages with the largest training corpora benefiting from adding a second language, and those with the smallest corpora (e.g. Indonesian, Italian or Japanese) suffering accuracy loss, likely because the training gets biased towards the second language.

Training bilingual models containing a significant number of non-overlapping treebank-dependent tags 
tends to have a positive effect. 
English and Spanish are two of the clearest examples of this.
As shown in Table \ref{table-shared-finetags}, which
shows a complete report of shared PoS tags for each pair of languages under the \emph{treebank-dependent} tags configuration,
English only shares 1 PoS tag with the rest of the corpora under the said configuration, 
except for Swedish, with up to 5 tags in common;
and the \emph{en-sv} model is the only one suffering a significant loss on the English test set.
Similar behavior is observed on Spanish: \emph{sv} (0), \emph{en} (1), \emph{ja} (10) and \emph{ko} (12) are the four languages with
fewest shared PoS tags, and those are the four that obtained a significant improvement on the 
Spanish evaluation; while with \emph{pt-br}, with 15 shared PoS tags, we lose accuracy. The validity of this hypothesis is reinforced by an experiment where we differentiate the universal tags by language by appending a language code to them (e.g. {\sc en\_noun} for an English noun). An overall improvement was observed with respect to the bilingual parsers with non-disjoint sets of features.

\begin{table}[hbtp]
\begin{center}
\small{
\tabcolsep=0.100cm
\begin{tabular}{|l|rrrrrrrrrr|}
\hline
&\bf\emph{de}&\bf\emph{en}&\bf\emph{es}
&\bf\emph{fr}&\bf\emph{id}&\bf\emph{it}
&\bf\emph{ja}&\bf\emph{ko}&\bf\emph{pt-br}
&\bf\emph{sv}\\
\hline
\bf\emph{de}&16&1&14&14&14&13&10&12&14&0\\
\bf\emph{en}&&45&1&1&1&1&1&1&1&5\\
\bf\emph{es}&&&24&14&14&13&10&12&15&0\\
\bf\emph{fr}&&&&14&14&13&10&12&14&0\\
\bf\emph{id}&&&&&14&13&10&12&14&0\\
\bf\emph{it}&&&&&&13&10&12&13&0\\
\bf\emph{ja}&&&&&&&763&10&10&0\\
\bf\emph{ko}&&&&&&&&20&12&0\\
\bf\emph{pt-br}&&&&&&&&&15&0\\
\bf\emph{sv}&&&&&&&&&&25\\
\hline
\end{tabular}}
\end{center}
\caption{\small{Shared language-specific tags between pairs of languages}}
\label{table-shared-finetags}
\end{table}

While all these experiments have been performed on sentences with gold PoS tags, preliminary experiments assuming predicted tags instead show analogous results: the absolute values of {\sc las} and {\sc uas} are slightly smaller across the board, but the behavior in relative terms is the same, and the bilingual models that improved over the monolingual baseline in the gold experiments keep doing so under this setting.

On the other hand, Table \ref{table-same-tags-sets-performance} shows the performance of
the monolingual and bilingual models under the \emph{universal tags only} configuration. 
The bilingual parsers are also able to keep an acceptable accuracy
with respect to the monolingual models, but significant losses
are much more prevalent than under the \emph{treebank-dependent tags} configuration.

Putting both tables together, our experiments clearly suggest that not only treebank-specific tags 
do not impair the training of bilingual models, but they are even beneficial, supporting the idea that using
partially treebank-dependent tagsets helps multilingual parsing. We hypothesize that this may be 
because complementing the universal information at the syntactic level with language-specific information at the lower levels (lexical and morphological) may help the parser identify specific constructions of one language that would not benefit from the knowledge learned from
the other, preventing it from trying to exploit spurious similarities between languages. This explanation is coherent 
with work on delexicalized parser transfer \cite{LynFosDraTou2014a} showing  that better results can be obtained using disparate 
languages than closely-related languages, as long as they have common syntactic constructions.
Thus, using universal PoS tags
to train multilingual parsers can be, surprisingly, counterproductive.

\subsection{Parsing code-switched sentences}

\begin{figure*}[t]
  \centering
  \includegraphics[width=16.4cm,clip]{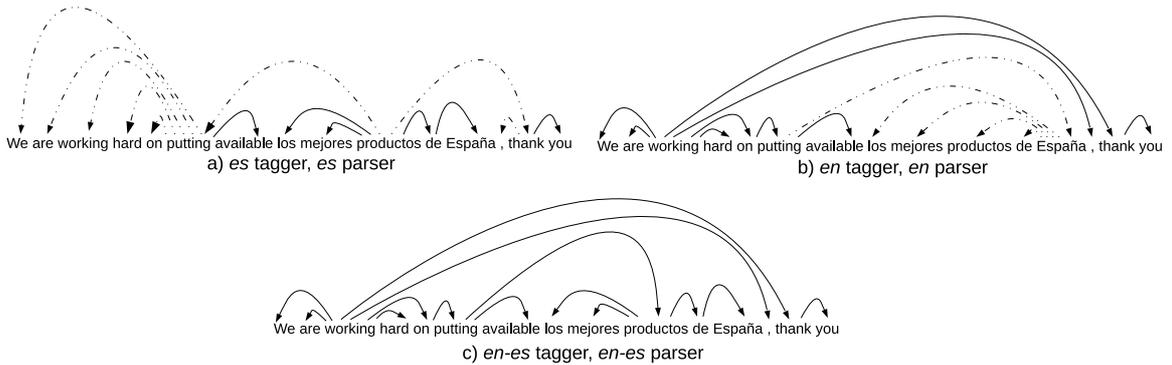}
  \caption{\small{Example  with the \emph{en}, \emph{es}, \emph{en-es} models. Dotted lines represent
  incorrectly-parsed dependencies. The corresponding English sentence is: \emph{\textquoteleft We are working hard on putting available the best products of Spain, thank you\textquoteright}}}
  \label{figure-dependency-parsing-example}
\end{figure*}

Our bilingual parsers also show robustness on texts exhibiting code-switching.
Unfortunately,
there are no syntactically annotated code-switching corpora, so we could not perform a formal evaluation.
We did perform informal tests, by running the Spanish-English bilingual parsers on some such sentences. 
We observed that they were able to parse the English and Spanish parts of the sentences much better than
monolingual models.
This required
training a bilingual tagger, which we did with the free distribution of the Stanford tagger \cite{TouMan2000a}; merging the Spanish and English corpora to train a combined bilingual tagger.
Under the \emph{universal tags only} configuration, the multilingual tagger 
obtained 98.00\% and 95.88\% over the monolingual test sets.
Using treebank-dependent tags instead, it obtained 97.19\% and 93.88\% over the monolingual test sets. Figure \ref{figure-dependency-parsing-example} 
shows an interesting example on how using bilingual parsers (and taggers) affects the parsing accuracy.

Table \ref{table-performance-code-switching-treebank} shows the performance on a tiny code-switching treebank built on top of ten normalized tweets.\footnote{The code-switching treebank follows the Universal Treebank v2.0 annotations. It can be obtained by asking any of the authors.} This confirms that monolingual pipelines perform poorly. Using a bilingual tagger helps improve the performance, thanks to accurate tags for both languages, but a bilingual parser is needed to push both {\sc las} and {\sc uas} up to state-of-the-art levels.

\begin{table}[]
\begin{center}
\small{
\begin{tabular}{|llrr|}
\hline
\bf Tagger&\bf Parser&\bf LAS&\bf UAS\\
\hline
\emph{en}&\emph{en}&37.82&44.23\\
\emph{es}&\emph{es}&27.56&41.03\\
\emph{en-es}&\emph{en}&66.03&78.85\\
\emph{en-es}&\emph{es}&67.95&77.56\\
\emph{en-es}&\emph{en-es}&\bf 87.18&\bf 92.31\\
\hline
\end{tabular}}
\end{center}
 \caption{\small{Performance on a code-switching treebank composed of 10 sentences.}}
\label{table-performance-code-switching-treebank}
\end{table}

\subsection{Adding more languages}

To show that our approach works when more languages are added, we created a quadrilingual parser using the romanic languages and the fine PoS tag set. 
The results ({\sc las}/{\sc uas}) on the monolingual sets were: 80.18/84.64 (\emph{es}), 79.11/84.29 (\emph{fr}), 82.16/86.15 (\emph{it}) and 84.45/86.80 (\emph{pt}).
In all cases, the performance is almost equivalent to the monolingual parser.

Noah's ARK group \cite{ammar2016one} has shown that this idea can be also adapted to universal parsing. Our models are a collection of weights learned from mixing harmonized treebanks, that accurately analyze sentences in any of the learned languages and where it is possible to take advantage of linguistic universals, but they are still dependent on language-specific word forms. Instead, \newcite{ammar2016one} rely on multilingual word clusters and multilingual word embeddings, learning a universal representation. They also support incorporating language-specific information (e.g. PoS tags) to keep learning language-specific behavior. 
To address syntactic differences between languages (e.g. noun-adjective  vs adjective-noun structure) they can inform the parser about the input language.

\section{Conclusions and future work}

To our knowledge, this is the first attempt to train purely bilingual parsers to 
analyze sentences irrespective of which of the two languages they are written in; as
existing work on training a parser on two languages \cite{SmiSmi04} focused
on using parallel corpora to transfer linguistic knowledge between languages.

Our results reflect that
bilingual parsers do not lose accuracy with respect to monolingual parsers
on their corresponding language, and can even outperform them,
especially if fine-grained tags are used. This shows 
that, thanks to universal dependencies and shared syntactic structures
across different languages, using treebank-dependent tag sets is not a drawback, but even
an advantage.

The applications include parsing sentences of different languages
with a single model, improving the accuracy of monolingual parsing with training sets from
other languages, and successfully parsing
sentences exhibiting code-switching.

As future work, our approach could benefit from simple domain adaptation techniques \cite{daume2009frustratingly}, to enrich the training set for a target language by incorporating data from a source language.

\section{Acknowledgments}

This research is supported by the  Ministerio de Econom\'{\i}a y Competitividad (FFI2014-51978-C2).
David Vilares is funded 
by the Ministerio de Educaci\'{o}n, Cultura y Deporte (FPU13/01180). Carlos G\'{o}mez-Rodr\'{\i}guez is funded by
an Oportunius program grant (Xunta de Galicia).
We thank Marcos Garcia for helping with the code-switching treebank. We also thank the reviewers for their comments and suggestions.

\bibliography{acl2015_vil_alo_gom}
\bibliographystyle{acl2016}

\end{document}